%% file: DFH.tex
\documentclass[11pt]{article}
\usepackage{fullpage}
\makeatletter
\long\def\@makecaption#1#2{
  \vskip 0.8ex
  \setbox\@tempboxa\hbox{\small {\bf #1:} #2}
  \parindent 1.5em  
  \dimen0=\hsize
  \advance\dimen0 by -3em
  \ifdim \wd\@tempboxa >\dimen0
  \hbox to \hsize{
    \parindent 0em
    \hfil 
    \parbox{\dimen0}{\def\baselinestretch{0.96}\small
      {\bf #1.} #2
    } 
    \hfil}
  \else \hbox to \hsize{\hfil \box\@tempboxa \hfil}
  \fi
}
\makeatother

\usepackage{times,enumitem,hyperref,booktabs,multicol,blindtext,url,parskip}
\usepackage{amsfonts}
\usepackage{amsmath}
\usepackage{amsthm}
\usepackage{amssymb}
\usepackage{caption}
\usepackage{microtype}
\usepackage{graphicx}
\usepackage{subcaption}
\usepackage{natbib}
\usepackage{fancyhdr}
\usepackage{color}
\usepackage{algorithm}
\usepackage{forloop}
\input{macros.tex}

\usepackage{multirow}
\usepackage{enumerate}
\usepackage{algpseudocode}
\usepackage{rotating}
\usepackage{makecell}
\usepackage{bm}

\newcommand{\heading}[1]{\noindent\textbf{#1}}

\def\circle#1{%
        \raisebox{.9pt}{\textcircled{\raisebox{-.9pt}{#1}}}%
}

\def\ie{\textit{i.e.}}

\def\etal{\textit{et~al.}}

\usepackage[capitalize]{cleveref}
\crefname{section}{Sec.}{Secs.}
\Crefname{section}{Section}{Sections}
\Crefname{table}{Table}{Tables}
\crefname{table}{Tab.}{Tabs.}

\def\shownotes{0}  
\ifnum\shownotes=1
\newcommand{\authnote}[2]{[#1: #2]}
\else
\newcommand{\authnote}[2]{}
\fi

\definecolor{mypink}{RGB}{219, 48, 122}

\usepackage{xcolor}

\hypersetup{
  colorlinks,
  citecolor=violet,
  linkcolor=red,
  urlcolor=blue}

\begin{document}

\abovedisplayskip=8pt plus0pt minus3pt
\belowdisplayskip=8pt plus0pt minus3pt

\begin{center}
  \vspace*{-1cm}
  {\LARGE Federated and Generalized Person Re-identification  \\
  \vspace{0.12cm}
  through Domain and Feature Hallucinating} \\
  \vspace{.4cm}
  {\Large Fengxiang Yang$^{\textcolor{mypink}{1}}$ ~~~~ Zhun Zhong$^{\textcolor{mypink}{2}}$ ~~~~ Zhiming Luo$^{\textcolor{mypink}{1}}$ \\
  \vspace{0.12cm}
  Shaozi Li$^{\textcolor{mypink}{1}}$ ~~~~ Nicu Sebe$^{\textcolor{mypink}{2}}$} \\
  \vspace{.4cm}
  \small{\textcolor{mypink}{1} Xiamen University} 
  \small{\textcolor{mypink}{2} University of Trento}
  
\end{center}

\input{abstract}

\input{intro}

\input{related}

\input{method}

\input{experiments}

\input{conclusion}

\newpage

\bibliography{egbib}
\bibliographystyle{iclr2022_conference}

\end{document}

%% file: macros.tex
\usepackage{amsthm}
\usepackage[english]{babel}
\usepackage[T1]{fontenc}
\usepackage{amsmath}
\usepackage{amssymb}
\usepackage{amsfonts}
\usepackage{multirow}
\usepackage{mathtools}
\usepackage{breqn}
\usepackage{bbm}
\usepackage{dsfont}
\usepackage{graphicx}
\usepackage{natbib}
\usepackage{cases}
\usepackage[colorinlistoftodos]{todonotes}

\newtheorem*{remark*}{Remark}

\newtheorem*{observation*}{Observation}

\numberwithin{equation}{section}

\newcommand{\Gnorm}[1]{{\left\vert\kern-0.25ex\left\vert\kern-0.25ex\left\vert #1 
		\right\vert\kern-0.25ex\right\vert\kern-0.25ex\right\vert}}
\newcommand{\gnorm}[1]{{\vert\kern-0.25ex\vert\kern-0.25ex\vert #1 
		\vert\kern-0.25ex\vert\kern-0.25ex\vert}}

%% file: abstract.tex
\begin{abstract}
In this paper, we study the problem of federated domain generalization (FedDG) for person re-identification (re-ID), which aims to learn a generalized model with multiple decentralized labeled source domains. An empirical method (FedAvg) trains local models individually and averages them to obtain the global model for further local fine-tuning or deploying in unseen target domains. One drawback of FedAvg is neglecting the data distributions of other clients during local training, making the local model overfit local data and producing a poorly-generalized global model. To solve this problem, we propose a novel method, called ``Domain and Feature Hallucinating (DFH)'', to produce diverse features for learning generalized local and global models. Specifically, after each model aggregation process, we share the Domain-level Feature Statistics (DFS) among different clients without violating data privacy. During local training, the DFS are used to synthesize novel domain statistics with the proposed domain hallucinating, which is achieved by re-weighting DFS with random weights. Then, we propose feature hallucinating to diversify local features by scaling and shifting them to the distribution of the obtained novel domain. The synthesized novel features retain the original pair-wise similarities, enabling us to utilize them to optimize the model in a supervised manner. Extensive experiments verify that the proposed DFH can effectively improve the generalization ability of the global model. Our method achieves the state-of-the-art performance for FedDG on four large-scale re-ID benchmarks.
\end{abstract}

\begin{figure}[!t]
    \centering
    \includegraphics[width=0.8\linewidth]{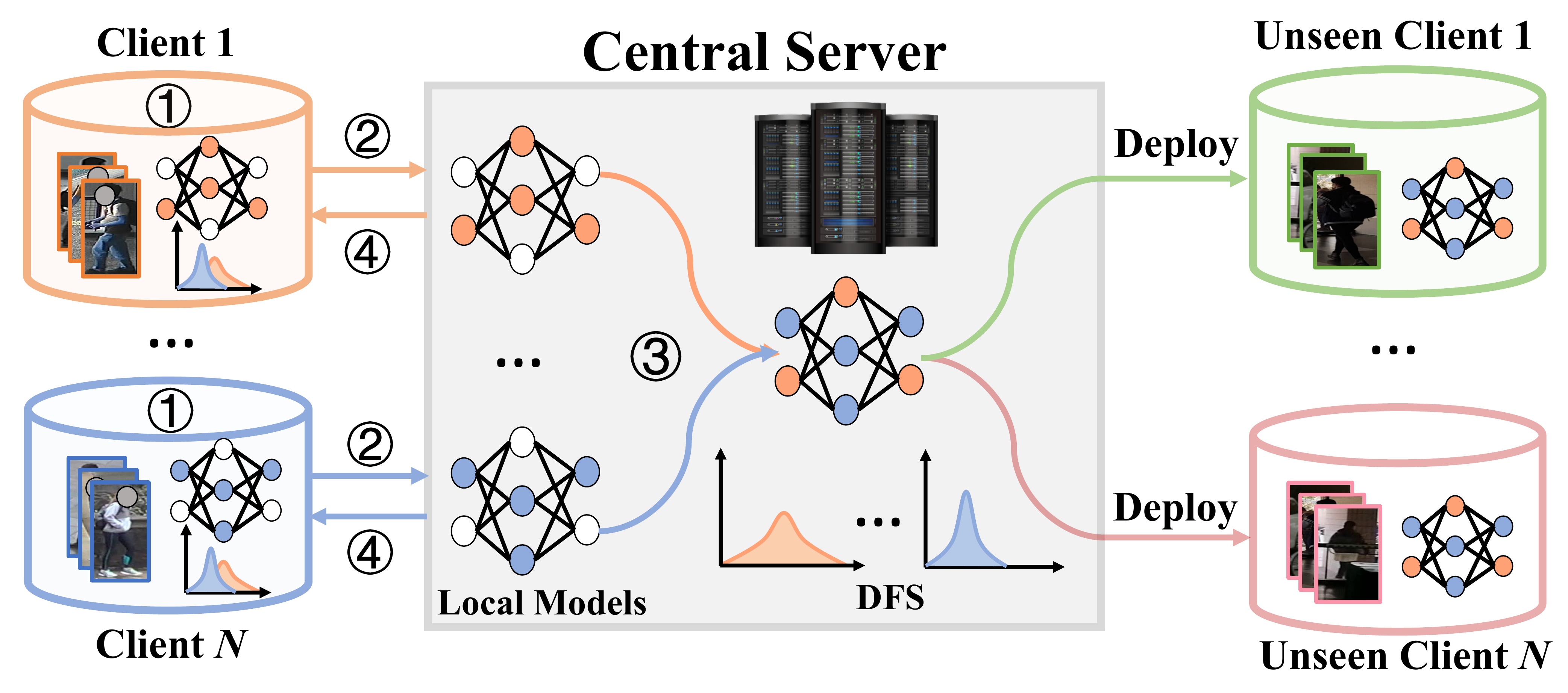}
    \vspace{-.2in}
    \caption{Illustration of our solution for FedDG re-ID. In the FedDG re-ID, each domain is considered as an individual client and we aim to train a generalized model with these decentralized data. In our solution, we share the domain-level feature statistics (DFS) of all clients to facilitate generalized re-ID training. The training process contains four steps: \circle1 local training with DFS downloaded from server, which synthesizes novel features with our proposed domain \& feature hallucinating (DFH) to improve the generalization of local models. \circle2 client-to-server updating, which updates DFS in server side. \circle3 server-side model aggregation. \circle4 redistributing global model and server-side DFS. By running \circle1 to \circle4 for several epochs, we can effectively improve the generalization of global model.}
    \vspace{-.1in}
    \label{fig:intro} 
\end{figure}

%% file: intro.tex
\section{Introduction}

Person re-identification (re-ID) aims at retrieving target pedestrians in a non-overlapped camera system. Thanks to the burgeoning development of deep learning, the accuracies of deep re-ID models have been boosted rapidly~\cite{sun2018beyond,wang2018learning,pami21reidsurvey}. Despite their promising performance in training domains, most of them can not generalize well to unseen domains. A plausible solution for this problem is domain generalization (DG), which aims to learn generalized models with source domains. However, existing advances in DG~\cite{zhao2021learning,song2019generalizable,zhou2020learning,dai2021generalizable,guo2020learning} usually demand the centralization of training data, which may bring privacy issues and thus limit the real-world applications.

Federated learning~\cite{mcmahan2017communication,liu2021feddg,yoon2021fedmix,li2021model,zhuang2020performance} is one way to alleviate the privacy issues, which utilizes 
multiple clients to jointly learn a model without exchanging their local data. Recently, Wu~\etal~\cite{wu2020decentralised} introduce a new paradigm, called federated domain generalized (FedDG) re-ID, to jointly solve the generalization and privacy problems. As illustrated in Fig.~\ref{fig:intro}, FedDG re-ID treats each domain as an individual client and aims to learn a generalized re-ID model without data transmission.~\cite{wu2020decentralised} achieves this goal with FedAvg~\cite{mcmahan2017communication}, which repeatedly iterates the local training  and the model aggregation processes. 
In addition, to avoid the model from overfitting on the local data,~\cite{wu2020decentralised} enforces the local model to mimic the predictions of the global model. However, this approach does not explicitly consider the data distribution of all clients during local training, hindering the further improvement of generalization for the global model.

In this paper, we introduce a new learning tactic for FedDG re-ID, which shares the domain-level feature statistics (DFS) of all domains without violating data privacy. For each domain, DFS are obtained by the mean and variance of identities.
Given the shared DFS, we propose a Domain and Feature Hallucinating (DFH) approach to generate diverse novel features, which are used to learn generalized models.
The process of our method is shown in Fig.~\ref{fig:intro}, which contains four steps. In step \circle1, we locally train the model of each client with the proposed DFH, where the local model and DFS are downloaded from the server. At each training iteration, we first propose Feature Hallucinating (FH) to diversify local features by scaling and shifting with the shared domain statistics of all clients. Then, to further diversify the local features, we synthesize novel domain statistics with the proposed Domain Hallucinating (DH), which is achieved by re-weighting DFS with randomly sampled weights. The novel statistics are also used in FH to generate novel features. The generated novel features retain the pair-wise similarities of the original ones, which can be utilized to improve generalization of local models with supervised learning. After local training, the local models and DFS are uploaded to the server (step \circle2) for updation. Next, we obtain the global model by averaging local models (step \circle3). Finally, the global model and DFS are redistributed to each client for the next epoch of  local training (step \circle4). With several iterations, we can obtain the final global model, which is deployed on unseen target domains. Our contributions are three-fold:

\begin{itemize}
    \item We introduce a new learning strategy for FedDG re-ID. By sharing the domain-level feature statistics (DFS) among domains, we are able to generate features of new distributions, without violating data privacy.
    
    \item We design a Feature Hallucinating (FH) method to generate novel features with the obtained domain statistics. The hallucinated features retain the pair-wise similarities of the original features, enabling us to use them to optimize the model in a supervised manner.
    
    \item We propose a Domain Hallucinating (DH) method to synthesize novel domain statistics, allowing the local models to see more diversified data distributions under the federated learning scenario.
    
\end{itemize}
\noindent Extensive experiments demonstrate the advantage of our learning strategy and DFH method in improving the generalization of the global model. Our method achieves state-of-the-art results on re-ID benchmarks for FedDG re-ID.

\vspace{-.1in}

%% file: related.tex
\section{Related Work}

\heading{Domain Generalization} Recent advances have paid more attention to training generalizable re-ID models, such as domain adaptation (DA)~\cite{zheng2021group,zhong2019invariance,yang2021joint,ge2020selfpaced}, and domain generalization (DG)~\cite{li2018learning,zhao2021learning,jin2020style,chattopadhyay2020learning}. DG optimizes re-ID models with several source domains and directly deploys the obtained model to target domain without further fine-tuning, which is a more applicable scenario than DA due to its low dependency on training data of target domain. Most DG methods focus on the close-set scenario~\cite{chattopadhyay2020learning,li2018learning,muandet2013domain,qiao2020learning}, considering that target domains have exactly the same IDs with the source domains and thus can hardly be applied to real-world applications. Different from them, DG for person re-ID is a more challenging problem because of the disjoint label space in the training and testing set. There are many algorithms designed for training domain generalized person re-ID~\cite{song2019generalizable,jin2020style,liao2020interpretable,zhao2021learning}. Jin~\etal~\cite{jin2020style} propose to disentangle features into ID-relevant and -irrelevant parts for subsequent feature stylization, reconstruction, and training. Liao~\etal~\cite{liao2020interpretable} improve the generalization of re-ID models by using their QAConv to compute novel sample-wise similarities. Zhao~\etal~\cite{zhao2021learning} utilize meta-learning to improve the generalization of models and integrate memory bank~\cite{zhong2019invariance} into the training process to address a series of problems raised by the open-set attribute~\cite{panareda2017open} in re-ID. These methods are successful, but they require combining several domains to achieve good results and may cause privacy problems when transferring data between different clients. In this paper, we consider a more practical and challenging scenario for DG in re-ID, which optimizes re-ID models with decentralized source data.

\heading{Federated Learning} In federated learning, training data are stored in several isolated local clients and can not be transferred to other clients to protect data privacy. McMahan~\etal~propose FedAvg~\cite{mcmahan2017communication} algorithm as the baseline method by averaging local models trained with local data and redistributing the averaged global model to local clients for further training. However, FedAvg may suffer from the non-IID or data heterogeneity problem~\cite{zhao2018federated} where the data distribution of local clients are different, leading to considerable performance degradation for the aggregated model. To solve this problem, FedProx~\cite{li2018federated} adds a regularization term to enforce the local model similar to the averaged global model. Recently, many researches~\cite{feng2020kd3a,liu2021feddg,ahmed2021unsupervised} extend the content of federated learning by applying it to various types of computer vision tasks, but most of them are designed for closed-set image classification or segmentation tasks, which have the same identities between training and testing sets. For the open-set~\cite{panareda2017open} re-ID problem, Zhuang~\etal~\cite{zhuang2020performance} explore the application of federated learning in re-ID and propose two settings called ``Federated-by-camera" and  ``Federated-by-identity" for the federated training of a single domain. Wu~\etal~\cite{wu2020decentralised} propose decentralized domain generalization for person re-ID to make the DG re-ID algorithms privacy-preserving. In this paper, we propose a novel DFH algorithm for FedDG re-ID problem and evaluate our model with large-scale benchmarks.

\begin{figure*}[!t]
    \centering
    \includegraphics[width=0.98\linewidth]{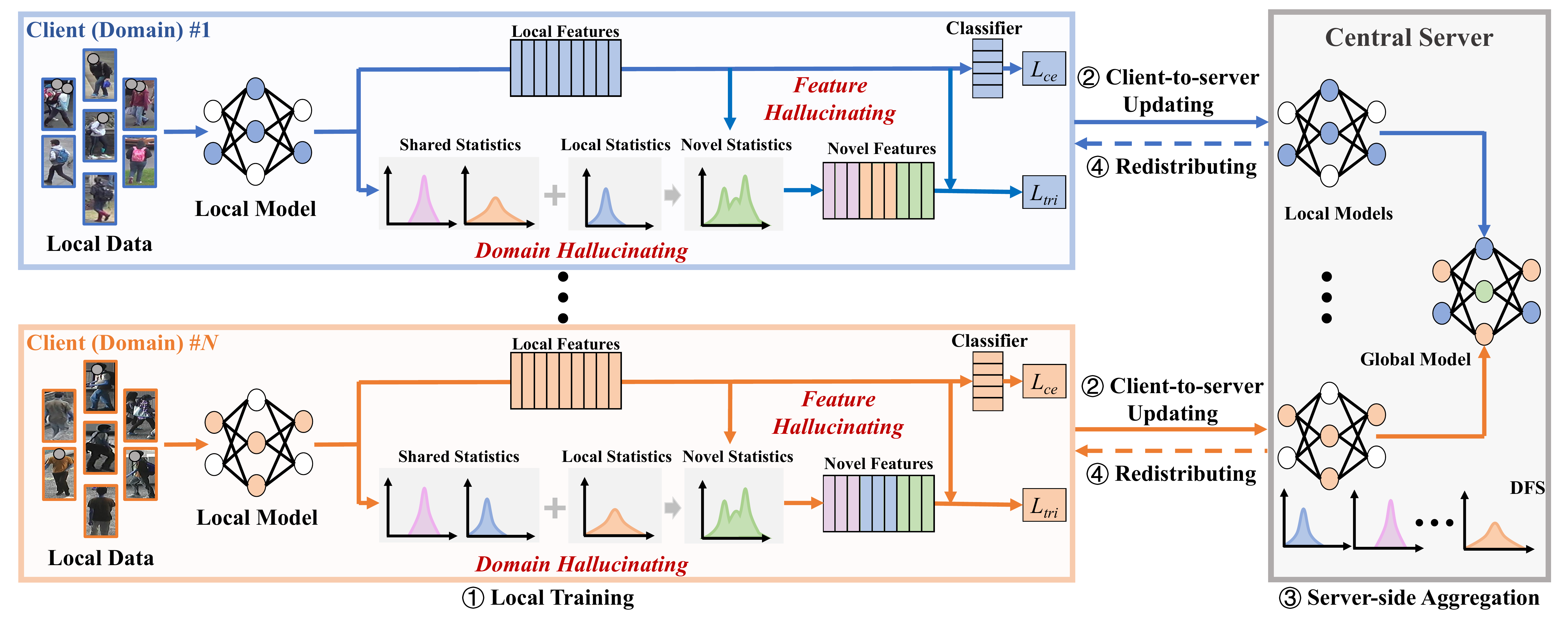}
    \vspace{-.15in}
    \caption{The overall framework of our proposed method. We propose to share domain-level feature statistics (DFS) for local training and the overall training can be divided into four steps. \circle1 Local training. \circle2 Client-to-server Updating. \circle3 Server-side aggregation. \circle4 Redistributing. The four steps are repeated for several steps to obtain final global model for evaluation. }
    \label{fig:fedreid} 
\end{figure*}

\vspace{-.1in}

%% file: method.tex
\section{Methodology}

\heading{Problem Definition} Suppose there are $N$ labeled source domains. For a given domain $i$ ($1 \leq i \leq N$), we have $M_i$ training images $\mathcal{X}_{i} = (x_1, x_2, ..., x_{M_i})$ and labels $\mathcal{Y}_{i} = (y_1, y_2, ..., y_{M_i})$ from $N_i$ IDs to train a local model $\phi_i$. The goal of FedDG re-ID is utilizing these decentralized data to optimize a well-generalized global re-ID model $\phi$. Since each domain is considered as an individual client, \textit{``client" and ``domain" have the same meaning in the context of FedDG re-ID}. The global model is then used to be evaluated in unseen domains.

\vspace{-.05in}
\subsection{Overview}
\vspace{-.05in}
\label{sec:overall}
The overall training process for one training epoch is shown in Fig.~\ref{fig:fedreid}, which contains four steps: \circle1 \textit{local training}, \circle2 \textit{client-to-server updating}, \circle3 \textit{server-side aggregation}, and \circle4 \textit{redistributing}. Before the optimization starts, we extract local features with corresponding local models to obtain mean and variance vectors for each identity ($\bm{\mu}_{i} \in \mathbb{R}^{N_{i} \times d}$ and $\bm{\sigma}_{i}^2 \in \mathbb{R}^{N_{i} \times d}$ are matrices with $N_i$ rows and $d$ columns, where $d$ is feature dimension and $N_{i}$ is the number of IDs for domain $i$). These ID-level feature statistics (IFS) are used to estimate domain-level feature statistics (DFS), which will be uploaded to server for public use. In step \circle1, we download DFS of all clients from the server to transform local batch features to other domains with the proposed feature hallucinating (FH). Moreover, we also enrich the statistical information by synthesizing novel domain statistics with our domain hallucinating (DH). Synthesized novel DFS, combined with DFS from other domains, are used to generate novel features with FH. These transformed features retain the pair-wise similarities of original ones and can be utilized to improve local generalization in a supervised manner. After executing local training for all clients, we upload local models and new DFS to the server to update the server-side DFS (step \circle2) and obtain the global model (step \circle3). In step \circle4, the global model and updated DFS are redistributed to each client to prepare for the next epoch of optimization. These four steps are repeated for several epochs to obtain the final model for evaluation.

\vspace{-.05in}
\subsection{Local Training}
\vspace{-.05in}
In this section, we introduce the strategy to estimate DFS for preparation, and demonstrate the local training process with our DFH.

\heading{Estimating DFS}. 
We model the features of a specific domain $D_{i}$ as a Gaussian distribution, which can be parameterized by mean ($\bm{\mu}_{D_i}$) and variance vectors ($\bm{\sigma}_{D_i}^2$).
A simple way to estimate them is averaging ID-level mean ($\bm{\mu}_{i}$) and variance vectors ($\bm{\sigma}_{i}^2$). This is because IFS roughly reflects the feature distribution of its corresponding client by showing how the class centroids are scattered in the feature space, and the intra-class variations for each ID.
However, the features of $D_{i}$ could be quite various, which is generally difficult to be represented by only two vectors $\bm{\mu}_{D_i}$ and $\bm{\sigma}_{D_i}^2$. To solve this drawback, we additionally treat the domain-level mean and variance vectors as another two Gaussian distributions, \ie, $\bm{\mu}_{D_i} \sim \mathcal{N}(\widehat{\bm{\mu}}_{i}, \widehat{\bm{\sigma}}_{i}^2 )$ and $\bm{\sigma}_{D_i}^2 \sim \mathcal{N}(\widetilde{\bm{\mu}}_{i}, \widetilde{\bm{\sigma}}_{i}^2 )$. We estimate them with IFS by using following equations:
\begin{equation}
    \begin{aligned}
        & \widehat{\bm{\mu}}_{i} = \frac{1}{N_i} \sum_{j=1}^{N_i} \bm{\mu}_{i, j}, \quad 
        (\widehat{\bm{\sigma}}_{i})^2 = \frac{1}{N_i} \sum_{j=1}^{N_i} \Big( \bm{\mu}_{i, j} - \widehat{\bm{\mu}}_{i} \Big)^2, \\
        & \widetilde{\bm{\mu}}_{i} = \frac{1}{N_i} \sum_{j=1}^{N_i} (\bm{\sigma}_{i, j}^2), \quad 
        (\widetilde{\bm{\sigma}}_{i})^2 = \frac{1}{N_i} \sum_{j=1}^{N_i} \Big( \bm{\sigma}_{i, j}^2 - \widetilde{\bm{\mu}}_{i} \Big)^2,
    \end{aligned}
    \label{eq:estimate}
\end{equation}
where $\bm{\mu}_{i,j}$ is the $j$-th row of IFS $\bm{\mu}_{i}$, denoting the mean vector of $j$-th ID in client $i$. $\bm{\sigma}_{i,j}^2$ has similar meaning. These four parameters ($\widetilde{\bm{\mu}}_{i}, \widetilde{\bm{\sigma}}_{i}^2, \widehat{\bm{\mu}}_{i}, \widehat{\bm{\sigma}}_{i}^2$) are the final DFS shared for local training. By sampling from these two Gaussian distributions (\ie, $\mathcal{N}(\widehat{\bm{\mu}}_{i}, \widehat{\bm{\sigma}}_{i}^2 )$ and $\mathcal{N}(\widetilde{\bm{\mu}}_{i}, \widetilde{\bm{\sigma}}_{i}^2 )$), we can generate diverse mean ($\bm{\mu}_{D_i}$) and variance ($\bm{\sigma}_{D_i}^2$) vectors to better describe domain $i$ while protecting the data privacy. We thereby send them to the server to improve local training with our feature hallucinating algorithm.

\heading{Feature Hallucinating}.
 During the local training of domain $i$, we first download DFS from the server and sample domain-level mean and variance vectors from them. We then sample a batch of data from $D_{i}$ to obtain local batch features $\mathbf{f}_i$, and transform them to another domain $k (k \neq i)$ with our feature hallucinating algorithm (FH), formulated as:
\begin{equation}
    \label{eq:fh}
    \mathbf{f}_{D_k} = \bm{\mu}_{D_k} + \bm{\sigma}_{D_k} BN(\mathbf{f}_i),
\end{equation}
where $\mathbf{f}_i \in \mathbb{R}^{N_b \times d}$ is a batch of feature vectors in $D_i$, extracted by the local model $\phi_i$. $BN$ is the pooling-5 batch-norm function of $\phi_i$, which transforms $\mathbf{f}_i$ to the standard Gaussian distribution. $\bm{\mu}_{D_k}$ and $\bm{\sigma}_{D_k}^2$ are sampled from the shared DFS (\ie, $\bm{\mu}_{D_k} \sim \mathcal{N}(\widehat{\bm{\mu}}_{k}, \widehat{\bm{\sigma}}_{k}^2 )$ and $\bm{\sigma}_{D_k}^2 \sim \mathcal{N}(\widetilde{\bm{\mu}}_{k}, \widetilde{\bm{\sigma}}_{k}^2$)) and will be replicated to the same size as $\mathbf{f}_i$ before computation. It should be noted that Eq.~\ref{eq:fh} is similar to batch-norm~\cite{ioffe2015batch}, where the affine parameters are replaced by the statistics in unseen domain $D_k$ ($\bm{\mu}_{D_k}$ and $\bm{\sigma}_{D_k}^2$) for scaling and shifting. Our FH enables the local model to improve local generalization without breaking the privacy constraint. To make a better use of the obtained DFS, we further enrich the local features with our domain hallucinating method for more generalized local training.

\heading{Domain Hallucinating}.
Our domain hallucinating (DH) aims at synthesizing novel domain-level mean and variance vectors by re-weighting vectors from other domains that are sampled from the shared DFS. Specifically, we sample domain weights from the Dirichlet distribution~\cite{shu2021open} and obtain the novel statistics $\bm{\mu}_{Novel}$ and $\bm{\sigma}_{Novel}^2$ with the following equations:
\begin{equation}
    \label{eq:dh}
    \begin{aligned}
        \mathbf{w} \sim & Dir(\bm{\alpha}), & \quad
        \bm{\mu}_{Novel} = \sum_{i=1}^{N} w_i \bm{\mu}_{D_i},& \quad
        \bm{\sigma}_{Novel}^2 = \sum_{i=1}^{N} w_i \bm{\sigma}_{D_i}^2,
    \end{aligned}
\end{equation}
where $\mathbf{w} \in \mathbb{R}^N$ is the domain weight vector sampled from Dirichlet distribution. The $i$-th element of $\mathbf{w}$ is $w_i$, which denotes the weight for client $i$. 
$\bm{\alpha}$ is the $N$-dim parameter for Dirichlet distribution. The synthesized novel domain statistics ($\bm{\mu}_{Novel} \in \mathbb{R}^d$ and $\bm{\sigma}_{Novel}^2 \in \mathbb{R}^d$ ) are different from all clients and can be used in our subsequent feature hallucinating. The domain hallucinating is motivated by MixUp~\cite{zhang2017mixup}, which shows that the beta-distribution is an effective to generate weights for two samples during the sampling of weights. Dirichlet distribution is a multivariate generalization of the beta-distribution~\cite{shu2021open}, enabling us to generate weights for multiple samples ($\geq 2$). Our setting assumes that there are multiple domains, we thus adopt Dirichlet distribution in our method.

\begin{algorithm}[!t]
    \caption{The Process of Local Training.}
    \label{alg:dh}
    \textbf{Inputs:} The total client number $N$. Total image number $N_{total}$. Local client $i$, its corresponding training data $\mathcal{X}_{i} = (x_1, x_2, ..., x_{N_i})$ and labels $\mathcal{Y}_{i} = (y_1, y_2, ..., y_{N_i})$. Local iteration number $iter$. \\
    \textbf{Outputs:} Local model $\phi_i$.
    \begin{algorithmic}[1]
        \State  \textcolor{gray}{// Domain Hallucinating.} 
        \Function{DH}{$\bm{\mu}_{D_1}$, $\bm{\sigma}_{D_1}^2$, ..., $\bm{\mu}_{D_N}$, $\bm{\sigma}_{D_N}^2$}
            \State Sample $\mathbf{w}$ from Dirichlet distribution;
            \State Synthesize novel domain statistics with Eq.~\ref{eq:dh};
            \State \textbf{Return} $\bm{\mu}_{Novel}$ and $\bm{\sigma}_{Novel}^2$;
        \EndFunction
        
        \State  \textcolor{gray}{// Local Training for Client $i$.} 
        \Function{LocalTrain}{$i$, $\widehat{\bm{\mu}}_{1}$, $\widehat{\bm{\sigma}}_{1}^2$, $\widetilde{\bm{\mu}}_{1}$, $\widetilde{\bm{\sigma}}_{1}^2$, ..., $\widehat{\bm{\mu}}_{N}$, $\widehat{\bm{\sigma}}_{N}^2$, $\widetilde{\bm{\mu}}_{N}$, $\widetilde{\bm{\sigma}}_{N}^2$}
            \For{$k$ in $N$}
                \State Sample $\bm{\mu}_{D_k}$, $\bm{\sigma}_{D_k}^2$ from $\mathcal{N}(\widehat{\bm{\mu}}_{i}, \widehat{\bm{\sigma}}_{i}^2)$ and $\mathcal{N}(\widetilde{\bm{\mu}}_{i}, \widetilde{\bm{\sigma}}_{i}^2)$;
            \EndFor
            \State ($\bm{\mu}_{Novel}$, $\bm{\sigma}_{Novel}^2$) $\leftarrow$ DH($\bm{\mu}_{D_1}$, $\bm{\sigma}_{D_1}^2, ..., \bm{\mu}_{D_N}$, $\bm{\sigma}_{D_N}^2$);
            
            \For{$iter\_num$ in $iter$}
                \State Sample a batch of training data from $\mathcal{X}_{i}$ and $\mathcal{Y}_{i}$;
                \State Extract batch features $\mathbf{F}_i$ with $\phi_i$;
                \State Transform $\mathbf{F}_i$ to $\mathbf{F}_{Novel}$ and $\mathbf{F}_{k} (k \neq i)$ with Eq.~\ref{eq:fh};
                \State Compute Eq.~\ref{eq:loss_all} to optimize $\phi_i$;
            \EndFor
            \State \textbf{Return} $\phi_i$;
        \EndFunction
    \end{algorithmic}
\end{algorithm}

\heading{Local Training with Synthesized Features}.
From Eq.~\ref{eq:fh}, we observe that FH can transform local features to another domain's distribution through scaling and shifting, which retains the pair-wise similarities of the original batch features. Therefore, we can utilize triplet loss~\cite{hermans2017defense} to optimize the local model with the synthesized novel feature and enhance the generalization of local models (optimization loss is not limited to triplet loss).  Concretely, we sample a training batch from ($\mathcal{X}_i$, $\mathcal{Y}_i$) of client $i$ with batch size $N_b$, then extract features $\mathbf{F}_i = \{ \mathbf{f}_{i,1}, \mathbf{f}_{i,2}, ..., \mathbf{f}_{i, N_b} \}$ with local model $\phi_i$. Each feature is then transformed to novel domain and other decentralized domains with Eq.~\ref{eq:fh} to obtain $\mathbf{F}_{Novel} = \{ \mathbf{f}_{Novel,1}, \mathbf{f}_{Novel,2}, ..., \mathbf{f}_{Novel, N_b} \}$ and $\mathbf{F}_{k} = \{ \mathbf{f}_{k,1}, \mathbf{f}_{k,2}, ..., \mathbf{f}_{k, N_b} \} (k \neq i)$. Thus, we optimize local model with the following triplet loss:
\begin{equation}
    \label{eq:trip}
    L_{tri} (\mathbf{F}) = \frac{1}{N_b} \sum_{n=1}^{N_b} \Big[ || \mathbf{f}_{n}^{+} - \mathbf{f}_{n} ||_2 
    - || \mathbf{f}_{n}^{-} - \mathbf{f}_{n} ||_2 + m \Big]_{+},
\end{equation}
where $\mathbf{F}$ can be $\mathbf{F}_{Novel}$ or $\mathbf{F}_{k} (k \neq i)$. $\mathbf{f}_{n}^{+}$ and $\mathbf{f}_{n}^{-}$ are hard positive and hard negative sample for anchor $\mathbf{f}_{n}$ within the batch. $m$ is margin, $N_b$ is the batch size, and $[\cdot]_{+}$ refers to $\max(\cdot, 0)$. Besides, we also use the original local features $\mathbf{F}_i$ for local optimization, formulated as:
\begin{equation}
    \label{eq:idloss}
    L_{ori}(\mathbf{F}_{i}, \mathcal{Y}_i) = L_{tri}(\mathbf{F}_{i}) + L_{ce}(FC(\mathbf{F}_{i}), \mathcal{Y}_i),
\end{equation}
where $FC$ is a fully connected classification layer, which transforms the input features $\mathbf{F}_{i}$ to logits. The logits and their corresponding ground-truth labels $\mathcal{Y}_i$ are used to compute cross entropy loss $L_{ce}$.

The total loss for local training is formulated as:
\begin{equation}
    \label{eq:loss_all}
    L_{local}(\mathbf{F}_{i}, \mathcal{Y}_i) = L_{ori}(\mathbf{F}_{i}, \mathcal{Y}_i) + \lambda \Big[ L_{tri}(\mathbf{F}_{Novel}) + \frac{1}{N-1} \sum_{k=1, k \neq i}^{N} L_{tri}(\mathbf{F}_{k}) \Big],
\end{equation}
where $\lambda$ is the balancing factor. The overall process of local training is shown in Alg.~\ref{alg:dh}.

\vspace{-.05in}
\subsection{Subsequent Learning}

\heading{Client-to-server Updating}. 
By running local training for all clients, we obtain well-generalized local models. Both local models and DFS (\ie, $\widehat{\bm{\mu}}_{i}$, $\widehat{\bm{\sigma}}_{i}^2$, $\widetilde{\bm{\mu}}_{i}$, $\widetilde{\bm{\sigma}}_{i}^2$ for the $i$-th domain) will be uploaded to the server. The server-side DFS will be updated by the local DFS to prepare for the next epoch of local training. Besides, the updated server-side DFS ensure that all the DFS used for local training are synchronous.

\heading{Server-side Aggregation}.
We aggregate all local models in server-side with weighted average. The weight for each model is decided by the relative ratio between the number of images in current client and all clients. In detail, the global model $\phi$ is obtained by:
\begin{equation}
    \label{eq:fedavg}
    \phi = \sum_{i=1}^{N} \frac{N_i}{N_{total}} \phi_{i},
\end{equation}
where $N_{total} = \sum_{i=1}^{N} N_i$ is the total number of images in all clients. 

\heading{Redistributing}.
The global model $\phi$, combined with server-side DFS (\ie, $\widehat{\bm{\mu}}_{1}$, $\widehat{\bm{\sigma}}_{1}^2$, $\widetilde{\bm{\mu}}_{1}$, $\widetilde{\bm{\sigma}}_{1}^2$, ..., $\widehat{\bm{\mu}}_{N}$, $\widehat{\bm{\sigma}}_{N}^2$, $\widetilde{\bm{\mu}}_{N}$, $\widetilde{\bm{\sigma}}_{N}^2$) are redistributed to all clients for the next epoch of federated learning. The aforementioned four steps are repeated for several epochs to obtain the final global model for evaluation. The overall process of FedDG re-ID is elaborated in Alg.~\ref{alg:fedreid}.

\begin{algorithm}[!t]
    \caption{The Overall Federated Learning.}
    \label{alg:fedreid}
    \textbf{Inputs:} Decentralized $N$ training domains and their labels ($\{ \mathcal{X}_{1}, \mathcal{Y}_{1}), ..., (\mathcal{X}_{N}, \mathcal{Y}_{N}) \}$). Initialized local models $\phi_{1}, \phi_{2}, ..., \phi_{N}$. Global training epoch $epoch$, batch-size $N_b$.. \\
    \textbf{Outputs:} Well-generalized global model $\phi$.
    \begin{algorithmic}[1]
        \State  \textcolor{gray}{// Initialize Server-side DFS.} 
        \For{$i$ in $N$}
            \State Obtain DFS for client-$i$ (\ie, $\widehat{\bm{\mu}}_{i}$, $\widehat{\bm{\sigma}}_{i}^2$, $\widetilde{\bm{\mu}}_{i}$, $\widetilde{\bm{\sigma}}_{i}^2$) and upload them to server;
        \EndFor
        \For{$m$ in $epoch$}
            \State  \textcolor{gray}{// Step 1: Local Training.} 
            \For{$k$ in $N$}
                \State  \textcolor{gray}{// Use downloaded DFS.}
                \State $\phi_k \leftarrow $ LocalTrain($k$, $\widehat{\bm{\mu}}_{1}$, $\widehat{\bm{\sigma}}_{1}^2$, $\widetilde{\bm{\mu}}_{1}$, $\widetilde{\bm{\sigma}}_{1}^2$, ..., $\widehat{\bm{\mu}}_{N}$, $\widehat{\bm{\sigma}}_{N}^2$, $\widetilde{\bm{\mu}}_{N}$, $\widetilde{\bm{\sigma}}_{N}^2$);
            \EndFor
            \State  \textcolor{gray}{// Step 2: Client-to-server Updating.} 
            \State Send local models to the server;
            \State Update server-side DFS;
            \State  \textcolor{gray}{// Step 3: Server-side Aggregation.} 
            \State Obtain global model $\phi$ with Eq.~\ref{eq:fedavg};
            \State  \textcolor{gray}{// Step 4: Redistributing.}
            \State  Redistribute $\phi$ and server-side DFS to local clients;
        \EndFor
        \State \textbf{Return} $\phi$;
    \end{algorithmic}
\end{algorithm}

\vspace{-.05in}
\subsection{Discussion on Data Privacy Constraint}

FedDG re-ID prohibits the transmission of images between different clients, which limits the application of most DG~\cite{dai2021generalizable,zhao2021learning,zhou2020learning,guo2020learning} algorithms. Generally speaking, the privacy constraint refers to no data transmission. However, recent works state that sharing averaged images~\cite{yoon2021fedmix}, domain-specific classifiers~\cite{wu2021collaborative} or intermediate feature distributions~\cite{luo2021no,yao2022federated} will not bring the leakage of data privacy since such weak information cannot be used to recover the original images. For example, FedMix~\cite{yoon2021fedmix} weakens the data-privacy constraint by using the averaged images from other clients for optimization. Wu~\etal~\cite{wu2021collaborative} attempt to handle the federated learning problem by jointly using the classifiers of each local model for optimization. Luo~\etal~\cite{luo2021no} propose CCVR to alleviate the heterogeneity in federated learning by sharing the feature statistics of each class for each domain. Despite their effectiveness in closed-set classification problem, most of them are not suitable in FedDG re-ID. This is because each domain has completely different IDs in FedDG re-ID. In this paper, we follow ~\cite{luo2021no,yao2022federated} and share domain statics between clients. Different from them, our method shares the global statistics in the pooling-5 feature space to synthesize novel features for optimization with re-parameterization trick~\cite{kingma2013auto}, offering a plausible way to handle the heterogeneity in the open-set FedDG re-ID.

%% file: experiments.tex
\section{Experiments}

\subsection{Experiment Setup}

\heading{Datasets}. In our experiments, we use four re-ID benchmarks, including Market-1501 (Market)~\cite{zheng2015scalable}, CUHK02~\cite{li2013locally}, MSMT-17 (MSMT)~\cite{wei2018person} and CUHK03~\cite{li2014deepreid}. Market contains $1,501$ IDs ($32,668$ images) taken by $6$ cameras, of which $750$ IDs ($12,936$ images) are used for training while the other $751$ IDs for evaluation. CUHK02 has $7,264$ bounding-boxes manually cropped from $1,816$ pedestrians. MSMT covers $126,441$ images from $4,101$ IDs captured by $15$ cameras. CUHK03 collects $28,193$ pedestrian images from $1,467$ IDs, where $1,367$ IDs are used for training and the rest for evaluation. Note that, DukeMTMC-ReID has been withdrawn by creators and thus is not used in this work.

\heading{Evaluation Protocol}. The overall performance is evaluated with the mean average precision (mAP) and rank-1 accuracy. We choose the final global model for evaluation.

\heading{Implementation Details}. We conduct all our experiments with ResNet-50~\cite{he2016deep}. During local training, we set learning rate as $1 \times 10^{-3}$, batch size $N_b=64$, local training iteration number $iter=200$, and the total server-client collaborative training epochs $epoch=40$. The learning rate will be multiplied with $0.5$ at the $20$-th and $30$-th epoch of training. The margin in Eq.~\ref{eq:trip} is set to $0.5$, $\lambda$ in Eq.~\ref{eq:loss_all} is set to $5$. We use random crop, random flip and random erasing~\cite{zhong2020random} as data augmentation during optimization and SGD is chosen as the optimizer. We set $\bm{\alpha}$ to all-one vector. All images are resized to $256 \times 128$. During testing, we extract $2048$-dim pooling-5 features for retrieval.

\begin{table*}[!t]
    \centering
    \scriptsize
    \renewcommand\arraystretch{1.2}
    \caption{Comparison with state-of-the-arts. We compare our methods with FedAvg~\cite{mcmahan2017communication}, FedPav~\cite{zhuang2020performance}, FedReID~\cite{wu2020decentralised} and SNR~\cite{jin2020style}. M: Market-1501, C2: CUHK02, C3: CUHK03, MS: MSMT-17.}
    \vspace{-.1in}
    \label{tab:sota}
    \begin{tabular}{c|c|cc|c|c|cc|c|c|cc} 
        \hline
        \multirow{2}{*}{Sources} & \multirow{2}{*}{Methods} & \multicolumn{2}{c}{Target: C2} \vline & \multirow{2}{*}{Sources} & \multirow{2}{*}{Methods} & \multicolumn{2}{c}{Target: M} \vline & \multirow{2}{*}{Sources} & \multirow{2}{*}{Methods} & \multicolumn{2}{c}{Target: C3} \\
        \cline{3-4}\cline{7-8}\cline{11-12}
         & & mAP & rank-1 & & & mAP & rank-1 & & & mAP & rank-1 \\
         \hline
         MS & \multirow{3}{*}{/} & 43.8 & 41.8 & MS & \multirow{3}{*}{/} & 23.3 & 47.5 & MS & \multirow{3}{*}{/} & 18.0 & 18.5 \\
         C3 & & 49.8 & 45.8 & C3 &  & 13.2 & 31.1 & C2 & & 21.6 & 22.5 \\
         M & & 42.4 & 37.9 & C2 & & 18.9 & 41.2 & M & & 10.2 & 11.2 \\
        \hline     
        \multirow{6}{*}{\makecell[c]{MS+\\C3+M}} & FedAvg & 61.3 & 58.6 & \multirow{6}{*}{\makecell[c]{MS+\\C3+C2}} & FedAvg & 24.6 & 48.3 & \multirow{6}{*}{\makecell[c]{MS+\\C2+M}} & FedAvg & 20.3 & 22.7 \\
        & FedPav & 62.6 & 59.8 & & FedPav & 25.4 & 49.4 & & FedPav & 22.5 & 24.3 \\
        & FedReID & 64.3 & 61.2 & & FedReID & 25.5 & 49.6 & & FedReID & 22.4 & 23.0 \\
        & SNR & 67.7 & 65.1 & & SNR & 28.3 & 53.2 & & SNR & 26.0 & 29.1 \\
        \cline{2-4}\cline{6-8}\cline{10-12}
        & DFH & \bf 71.7 & \bf 69.5 & & DFH & \bf 31.3 & \bf 56.5 & & DFH & \bf 27.2 & \bf 30.5 \\
        & SNR+DFH & \bf 72.4 & \bf 70.2 & & SNR+DFH & \bf 33.2 & \bf 58.8 & & SNR+DFH & \bf 31.1 & \bf 33.5 \\
        \hline
    \end{tabular}
\end{table*}

\subsection{Comparison with State-of-the-arts}
We alternately use CUHK02, Market, and CUHK03 as the target client and the other datasets as decentralized source clients for federated learning. FedAvg~\cite{mcmahan2017communication}, FedPav~\cite{zhuang2020performance}, FedReID~\cite{wu2020decentralised} and SNR~\cite{jin2020style} are the state-of-the-arts used in our experiments for comparison. FedAvg requires the classifier in client-side and server-side have the same number of outputs. Therefore, it will collect the total number of IDs in all source clients as the output of classifiers for all models. Differently, FedPav allows local models to have classifiers with a different number of outputs, and only transmits the feature extractor of models for aggregation. FedReID improves the accuracies of FedDG re-ID by using knowledge distillation between the aggregated model and the local model. SNR is a single domain generalization method, which modifies the network structure of ResNet-50 by adding SNR module for feature disentanglement and better training. We use SNR under the federated constraint for fair comparison. We report all results in Tab.~\ref{tab:sota} and obtain the following conclusions: (1) Our method achieves the state-of-the-art performance on FedDG re-ID. Concretely, we achieve \textbf{mAP=71.7\%} and \textbf{rank-1=69.5\%} when testing on CUHK02, \textbf{mAP=31.3\%} and  \textbf{rank-1=56.5\%} when testing on Market, and \textbf{mAP=27.2\%} and  \textbf{rank-1=30.5\%} when testing on CUHK03. These results outperform FedPav by 9.1\%, 5.9\%, and 4.7\% mAP scores on the three benchmarks respectively. Besides, our method also outperforms SNR~\cite{jin2020style} by 4.0\%, 3.0\%, and 1.2\% mAP scores on the three datasets without changing the network's structure, indicating the effectiveness of our model. (2) Federated learning with multiple domains achieves better generalization results than using a single domain. We take the results on CUHK02 for example, the models trained with MSMT, CUHK03, and Market respectively achieve 43.8\%, 49.8\%, and 42.4\% mAP scores on the test set of CUHK02, which are far less than the results of multi-source FedDG re-ID methods like FedPav and ours. The results demonstrate that although federated learning does not combine all training data for optimization, it can still achieve better generalization than models with one domain for training. (3) Our method can be flexibly integrated into some DG methods like SNR~\cite{jin2020style}, and help them improve federated re-ID accuracies. As shown in Tab.~\ref{tab:sota}, SNR achieves 28.3\% mAP score on Market1501. After adding our DFH into the training process, the mAP is increased by 4.9\%, which indicates the effectiveness of DFH on SNR under federated constraint. Similar results can also be found in the evaluation results of other benchmarks.

\begin{table}[!t]
    \centering
    \small
    \renewcommand\arraystretch{1.2}
    \caption{Ablation study of our method. FH: Feature Hallucinating, DM: Dirichlet MixUp~\cite{shu2021open}, FM: Feature-level MixUp~\cite{zhang2017mixup}, DH: Our Domain Hallucinating.}
    \vspace{-.1in}
    \label{tab:ablation}
    \begin{tabular}{c|cccc|cc|cc} 
        \hline
        \multirow{2}{*}{No.} & \multicolumn{4}{c}{Attributes} \vline & \multicolumn{2}{c}{\makecell[c]{MS+C2 \\ +C3$\rightarrow$M}} \vline & \multicolumn{2}{c}{\makecell[c]{MS+M \\ +C2$\rightarrow$C3}} \\
        \cline{2-9}
         & FH & DM & FM & DH & mAP & rank-1 & mAP & rank-1 \\
         \hline
         1 & $\times$ & $\times$ & $\times$ & $\times$ & 25.4 & 49.4 & 22.5 & 24.3 \\
         2 & $\checkmark$ & $\times$ & $\times$ & $\times$ & 28.4 & 51.7 & 24.1 & 25.7 \\
         3 & $\checkmark$ & $\checkmark$ & $\times$ & $\times$ & 29.1 & 52.2 & 25.6 & 28.2 \\
         4 & $\checkmark$ & $\times$ & $\checkmark$ & $\times$ & 27.8 & 50.5 & 23.6 & 25.3 \\
         5 & $\checkmark$ & $\times$ & $\times$ & $\checkmark$ & \bf 31.3 & \bf 56.5 & \bf 27.2 & \bf 30.5 \\ 
        \hline
    \end{tabular}
    \vspace{-.15in}
\end{table}

\subsection{Ablation Study}
We evaluate the effectiveness of DH, FH, and compare our DFH with other feature-level MixUp methods under the federated learning scenario. In detail, we conduct experiments that (1) do not use DH and FH; (2) do not synthesize novel domain with DH, but use FH to enable local model to see other clients'  feature distributions; (3) use Dirichlet-MixUp~\cite{shu2021open} on transformed features to generate novel features; (4) use Feature-level MixUp~\cite{zhang2017mixup} to generate novel features and sample weights from beta distribution; (5) use DFH; All results are reported in Tab.~\ref{tab:ablation}.

\heading{The Effectiveness of Feature Hallucinating}.
From \textit{No.~1} and \textit{No.~2} of Tab.~\ref{tab:ablation}, we notice that FH enables the local model to see the feature distributions of other domains, leading to the improvement of re-ID accuracies. In detail, for experiments utilizing Market as the target domain, \textit{we improve the mAP from 25.4\% to 28.4\%}. Similar results can also be found on experiments when evaluated on CUHK03, which elaborate the effectiveness of FH for federated learning.

\heading{The Effectiveness of Domain Hallucinating}.
From \textit{No.~2} and \textit{No.~5} of Tab.~\ref{tab:ablation}, we conclude that synthesizing novel domain statistics is also beneficial to improving the generalization of the re-ID model. \textit{After applying both FH and DH to the FedDG re-ID, the mAP scores are further improved from 28.4\% to 31.3\% on Market}. We can also receive similar results on CUHK03, which suggest the importance of synthesizing novel domain statistics during federated learning.

\heading{Comparison between DH and Other Feature-level MixUp Methods}.
To clarify the differences between our DH and other feature MixUp methods, we replace the DH with another two methods: Dirichlet-Mixup~\cite{shu2021open} and Feature-level MixUp~\cite{zhang2017mixup}. The former approach samples weights from Dirichlet distribution and generates novel features with the hallucinated features, which is plausible when there are more than two features to mixup. The latter one follows the original MixUp and samples weights from beta-distribution. By comparing \textit{No.~3} and \textit{No.~4}, we conclude that for FedDG re-ID problem with multiple domains, it is better to sample weights from Dirichlet distribution. Moreover, by comparing \textit{No.~3}, \textit{No.~5}, we notice that our DH achieves better results than Dirichlet-Mixup. Therefore, our DFH can bring more diverse local features during local training and can effectively improve local generalization.

\begin{table}[!t]
    \centering
    \small
    \renewcommand\arraystretch{1.2}
    \caption{Effectiveness of training with less source clients.}
    \vspace{-.1in}
    \label{tab:multisrc}
    \begin{tabular}{c|cc|cc|cc|cc} 
        \hline
        \multirow{2}{*}{Methods} & \multicolumn{2}{c}{MS+C3$\rightarrow$M} \vline & \multicolumn{2}{c}{MS+C2$\rightarrow$M} \vline & \multicolumn{2}{c}{MS+M$\rightarrow$C3} \vline & \multicolumn{2}{c}{MS+C2$\rightarrow$C3} \\
        \cline{2-9}
         & mAP & rank-1 & mAP & rank-1 & mAP & rank-1 & mAP & rank-1 \\
         \hline
         FedPav & 24.6 & 45.3 & 24.3 & 47.8 & 18.9 & 19.5 & 22.8 & 26.5 \\
         SNR & 26.3 & 49.7 & 28.2 & 53.6 & 20.9 & 22.0 & 23.2 & 22.4 \\
         Ours & \bf 29.8 & \bf 54.9 & \bf 29.9 & \bf 55.1 & \bf 23.5 & \bf 25.2 & \bf 25.0 & \bf 28.7 \\
        \hline
    \end{tabular}
    \vspace{-.1in}
\end{table}

\subsection{Sensitivity Analysis}

\heading{Sensitivity to the Number of Clients}.
Previous experiments are all conducted with three decentralized source domains. To explore the effectiveness of our method when deployed on fewer clients, we report experiments with only two source decentralized clients in Tab.~\ref{tab:multisrc}. By observing the results, we notice that our approach can improve the generalization of re-ID models with less (two) decentralized domains. Concretely, DFH achieves higher re-ID accuracies than FedPav~\cite{zhuang2020performance} and SNR~\cite{jin2020style} on ``MS+C3$\rightarrow$M", ``MS+C2$\rightarrow$M", ``MS+M$\rightarrow$C3" and ``MS+C2$\rightarrow$C3". For the comparison with FedPav, DFH can consistently improve the re-ID accuracies for about 3-5\% mAP scores. The experiments of SNR report around 2\% improvements on mAP, indicating that our method is also effective when there are fewer source domains.

\heading{Sensitivity to $\bm{\alpha}$}.
We explore the effect of using different $\bm{\alpha}$ when sampling domain weight from the Dirichlet distribution during DH. Specifically, $\bm{\alpha}$ is a $N$ dimensional vector where each element is related to a domain. We change one of the elements in $\bm{\alpha}$ from 1 to 2 and remain the others fixed for ``MS+M+C2$\rightarrow$C3" and ``MS+C3+C2$\rightarrow$M" to see how it affects the final performance. The results are reported in Tab.~\ref{tab:SA}. We find that changing one of the elements in $\bm{\alpha}$ does not have a significant effect on the final results. In real-world applications, we recommend setting $\bm{\alpha}$ to all-one vector for simplicity.

\begin{table}[!t]
    \centering
    \renewcommand\arraystretch{1.2}
    \caption{Sensitivity analysis of $\bm{\alpha}$. }
    \vspace{-.1in}
    \label{tab:SA}
    \begin{tabular}{c|c|c|cc|c|c|c|cc} 
        \hline
        \multicolumn{3}{c}{$\bm{\alpha}$} \vline & \multicolumn{2}{c}{Target: C3} \vline & \multicolumn{3}{c}{$\bm{\alpha}$} \vline & \multicolumn{2}{c}{Target: M} \\
        \hline
          MS & M & C2 & mAP & rank-1 & MS & C3 & C2 & mAP & rank-1 \\
         \hline
         1 & 1 & 1 & \bf 27.2 & \bf 30.5 & 1 & 1 & 1 & \bf 31.3 & \bf 56.5 \\
         1 & 2 & 1 & 26.9 & 25.5 & 1 & 2 & 1 & 30.7 & 55.3 \\
         2 & 1 & 1 & 25.8 & 28.7 & 2 & 1 & 1 & 29.8 & 54.6 \\
         1 & 1 & 2 & 26.4 & 30.2 & 1 & 1 & 2 & 30.2 & 54.8 \\
        \hline
    \end{tabular}
    \vspace{-.1in}
\end{table}

\begin{figure}[!t]
    \centering
    \includegraphics[width=0.65\linewidth]{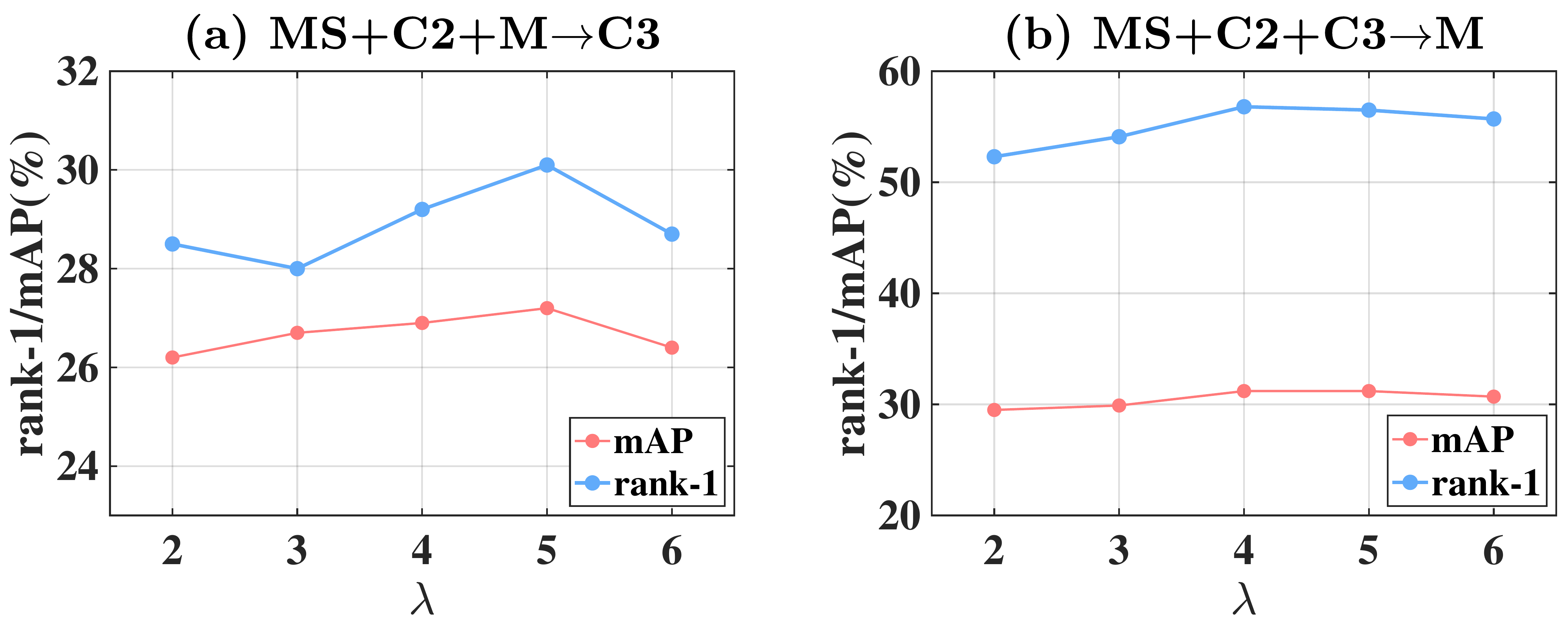}
    \vspace{-.05in}
    \caption{Sensitivity analysis of $\lambda$. We change the $\lambda$ from 2 to 6 with a step size of 1 to see the importance of using transformed features during optimization.}
    \label{fig:lam} 
    \vspace{-.15in}
\end{figure}

\heading{Sensitivity to $\lambda$}.
We investigate the effect of $\lambda$ by changing its value from 2 to 6 with a step size of 1 and apply them in two experiments (``MS+C2+M$\rightarrow$C3" and ``MS+C2+C3$\rightarrow$M"). A higher $\lambda$ means paying more attention to the transformed features during optimization. From Fig.~\ref{fig:lam}, we can draw a conclusion that focusing on the transformed features during optimization and using larger $\lambda$ is beneficial to improving the generalization of models due to the novel information included in the transformed features. But, the improvement of re-ID accuracies begins to become small when $\lambda > 5$. Therefore, we choose $\lambda=5$ for FedDG re-ID tasks.

\subsection{Loss Function}. We want to remind the readers that the hallucinated features can not be optimized with cross-entropy loss and local classifiers. As shown in Tab.~\ref{tab:trip}, the results of using cross-entropy loss for optimization is far less than the triplet version. That is because FH generates features through affine transformation, which will change the neighbors of the original ones. Since the transformed features and the original ones have different neighbors, making them to be similar with cross-entropy loss will back-propagate false information to the network and reduce the discrimination of the model.

\begin{table}
    \centering
    \small
    \renewcommand\arraystretch{1.2}
    \caption{The Necessity of Using Triplet Loss.}
    \vspace{-.1in}
    \label{tab:trip}
    \begin{tabular}{c|cc|cc} 
        \hline
        \multirow{2}{*}{Loss Function} & \multicolumn{2}{c}{MS+M+C2$\rightarrow$C3} \vline & \multicolumn{2}{c}{MS+C2+C3$\rightarrow$M}\\
        \cline{2-5}
         & mAP & rank-1 & mAP & rank-1 \\
         \hline
         Cross-Entropy & 20.4 & 23.8 & 26.3 & 50.6 \\
         Triplet & \bf 27.2 & \bf 30.5 & \bf 31.3 & \bf 56.5 \\
        \hline
    \end{tabular}
    \vspace{-.1in}
\end{table}

\subsection{Visualization}

\begin{figure}
    \centering
    \includegraphics[width=0.8\linewidth]{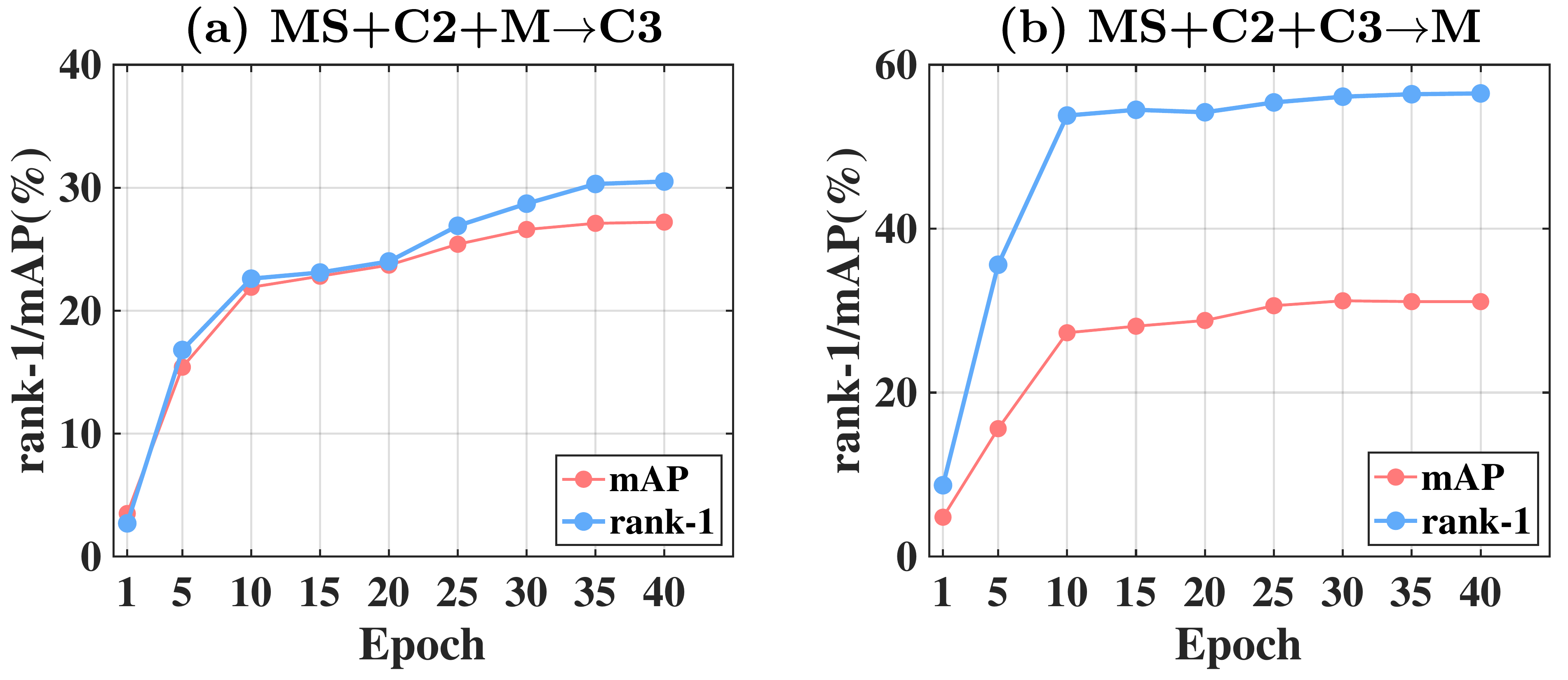}
    \caption{Convergence of DFH.}
    \label{fig:convergence} 
\end{figure}

 \heading{Convergence of DFH}. We evaluate the mAP and rank-1 scores of the  averaged model every 5 epochs to check the convergence of our DFH in the whole 40 epochs. As shown in Fig.~\ref{fig:convergence}, the mAP and rank-1 scores become steady at the 35-th epoch in ``MS+C2+M$\rightarrow$C3", while ``MS+C2+C3$\rightarrow$M" converged at the 10-th epoch. 
 
\heading{Results of Restored Images with D2D}.
Transferring the raw features to different clients may cause some information leakage because researchers have developed algorithms like D2D~\cite{yin2020dreaming} to restore the original images from the frozen model. It, therefore, raises questions whether sharing the domain-level feature statistics can preserve data privacy. We conduct experiments to restore images with the trained Market model and D2D.

\begin{figure}
    \centering
    \includegraphics[width=0.7\linewidth]{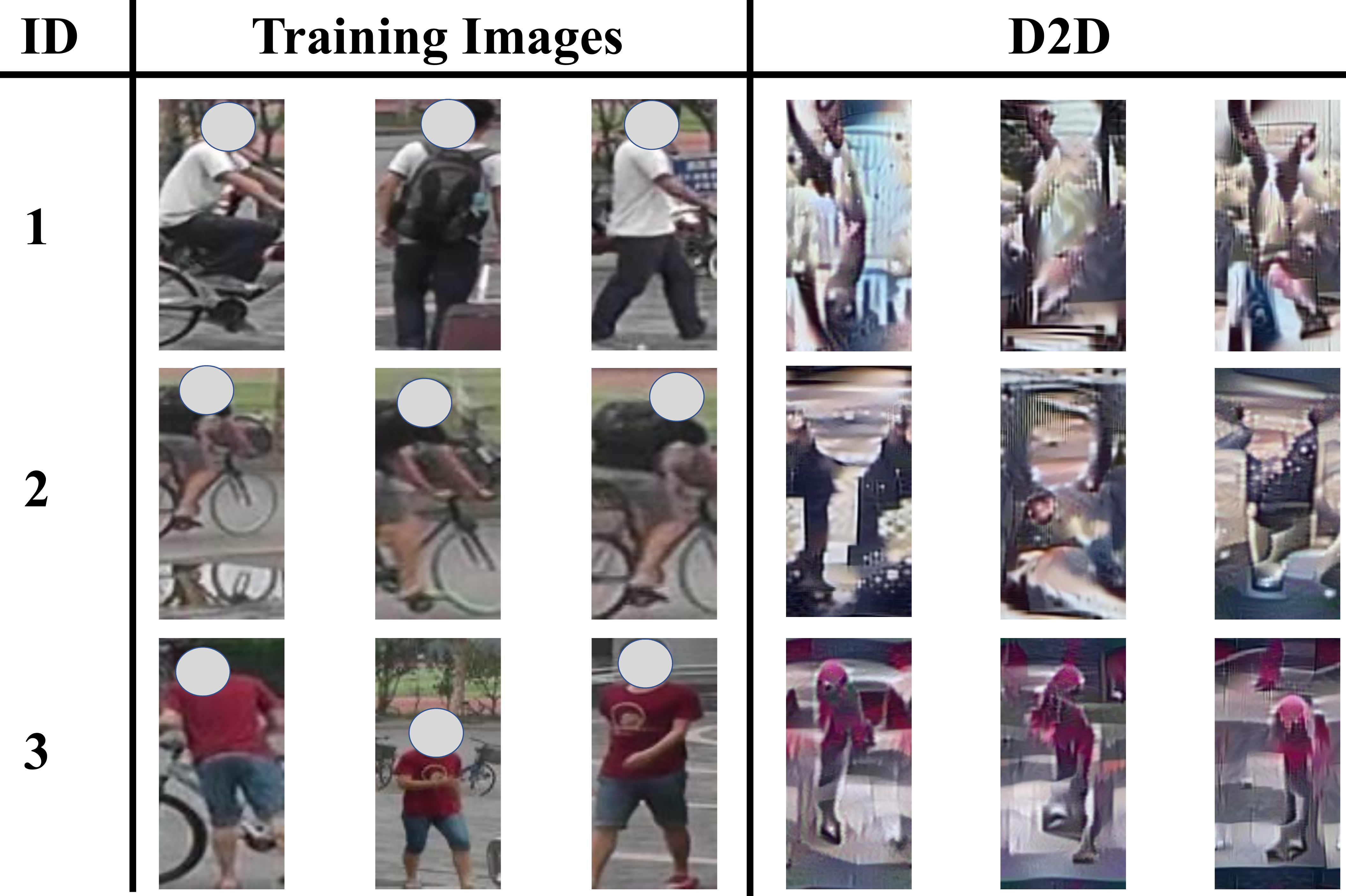}
    \caption{Restored Images with D2D.}
    \label{fig:d2d} 
\end{figure}

The shared DFS are parameters of two Gaussians, which should be first sampled to obtain the domain-level mean and variance. The receiver cannot even know the number of IDs from DFS and will cause some problems for D2D to recover the raw images. Therefore, we use IFS to restore the raw images and the mean vectors for each ID in the Market are used to initialize the classifier during the optimization. If IFS can successfully protect data privacy, the DFS will also satisfy privacy constraint. The results are shown in Fig.~\ref{fig:d2d}. It is obvious that the restored images are blurry and we cannot recognize the object in them. Based on the results, we claim that transferring DFS will not violate data privacy constraint.

\begin{figure}[!t]
    \centering
    \includegraphics[width=0.95\linewidth]{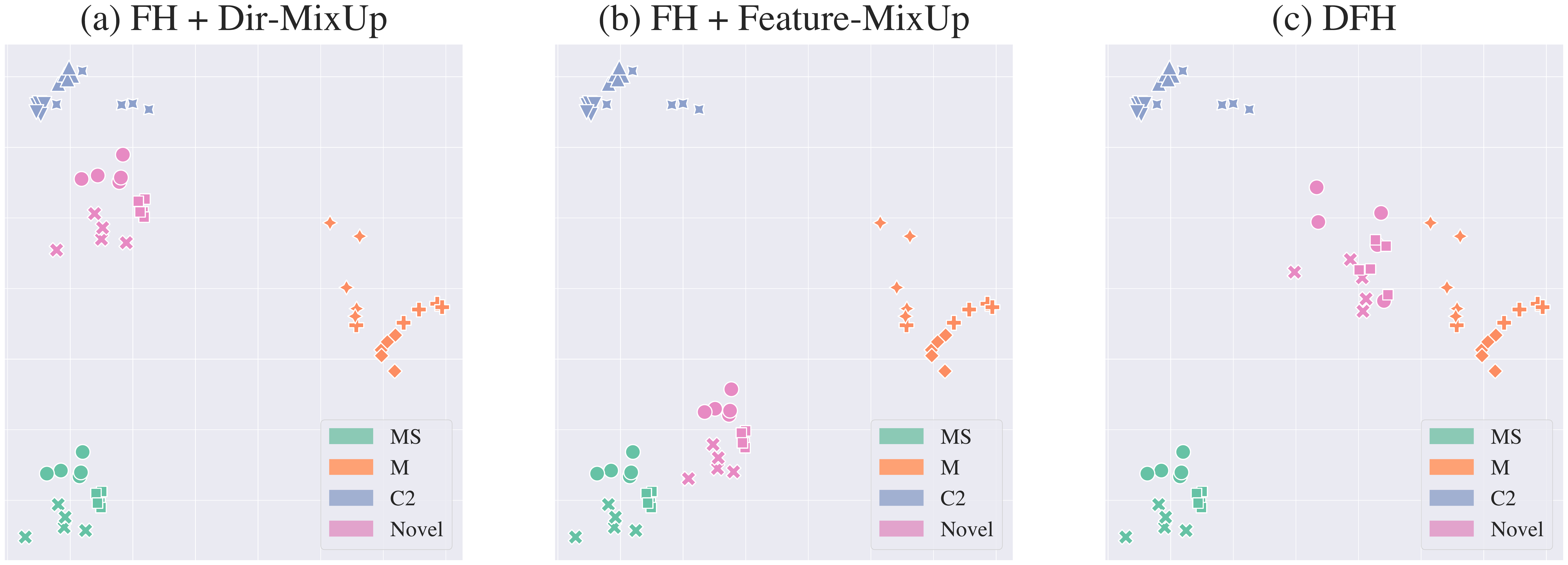}
    \vspace{-.1in}
    \caption{Visualization of novel features generated by FH+Dir-MixUp, FH+Feature-MixUp, and DFH. (a): FH+Dir-MixUp, which retains nearly the same pair-wise similarities of local features. (b): FH+Dir-MixUp, which can only interpolate the transition between each two domains. (c): Our DFH.}
    \label{fig:vis} 
    \vspace{-.25in}
\end{figure}

\heading{Visualization of the Hallucinated Features}. 
To gain a better understanding of how DFH and other feature mixup methods work in FedDG re-ID task, we conduct a visualization experiment, which chooses MSMT, CUHK02, and Market as the decentralized domains. We collect images from randomly selected three identities for each domain and extract their features with the final global model. Then, we respectively use Dirichlet-MixUp~\cite{shu2021open}, Feature MixUp~\cite{zhang2017mixup}, and our DFH to generate novel features from MSMT. All features are visualized with \emph{t}-SNE~\cite{van2008visualizing} in Fig.~\ref{fig:vis}. We use different colors to denote different domains while different shapes to indicate identities. In Fig.~\ref{fig:vis}, we have three observations. \textbf{(1) All three methods try to improve the model's generalization by synthesizing novel features in the transition among different domains}. Specifically, the transformed features are located in the gap of three source domains, which enables the local model to see more diverse samples and avoids overfitting by smoothing the transition between different domains. \textbf{(2) We should use Dirichlet-MixUp for generalization tasks with multiple domains}. As mentioned in~\cite{shu2021open}, the classical MixUp only mix features from two domains with weights sampled from beta-distribution, which can not generate diverse features for problems with multiple domains. The comparison between Fig.~\ref{fig:vis}-(a) and (b) also demonstrates the necessity of using Dirichlet-MixUp. \textbf{(3) Our DFH can further improve the generalization of models by enlarging the intra-class variations}. In Fig.~\ref{fig:vis}-(a), the transformed features nearly retain the same pair-wise similarities of the original features, while our DFH in Fig.~\ref{fig:vis}-(c) can generate more diverse features. This is because DFH generates new features by applying affine transformation with synthesized novel feature statistics, which avoid applying MixUp in feature level and improve the diversity of novel features.

%% file: conclusion.tex
\section{Conclusion}
In this paper, we share domain-level feature statistics (DFS) of different clients and propose domain \& feature hallucinating (DFH) to improve the generalization of FedDG re-ID. During the local training of each client, we estimate the feature distribution of local clients with the shared DFS. Then, we synthesize novel domain statistics by interpolating shared DFS with randomly sampled domain weights. The novel domain statistics are utilized by our feature hallucinating to transform local features to novel distributions, enabling the local model to see as diverse samples as possible and improving the generalization of local and global models. Extensive experiments demonstrate the effectiveness of our method. 

%% file: DFH.bbl
\begin{thebibliography}{45}
\providecommand{\natexlab}[1]{#1}
\providecommand{\url}[1]{\texttt{#1}}
\expandafter\ifx\csname urlstyle\endcsname\relax
  \providecommand{\doi}[1]{doi: #1}\else
  \providecommand{\doi}{doi: \begingroup \urlstyle{rm}\Url}\fi

\bibitem[Ahmed et~al.(2021)Ahmed, Raychaudhuri, Paul, Oymak, and
  Roy-Chowdhury]{ahmed2021unsupervised}
Sk~Miraj Ahmed, Dripta~S Raychaudhuri, Sujoy Paul, Samet Oymak, and Amit~K
  Roy-Chowdhury.
\newblock Unsupervised multi-source domain adaptation without access to source
  data.
\newblock In \emph{Proceedings of the IEEE/CVF Conference on Computer Vision
  and Pattern Recognition}, 2021.

\bibitem[Chattopadhyay et~al.(2020)Chattopadhyay, Balaji, and
  Hoffman]{chattopadhyay2020learning}
Prithvijit Chattopadhyay, Yogesh Balaji, and Judy Hoffman.
\newblock Learning to balance specificity and invariance for in and out of
  domain generalization.
\newblock In \emph{European Conference on Computer Vision}, 2020.

\bibitem[Dai et~al.(2021)Dai, Li, Liu, Tong, and Duan]{dai2021generalizable}
Yongxing Dai, Xiaotong Li, Jun Liu, Zekun Tong, and Ling-Yu Duan.
\newblock Generalizable person re-identification with relevance-aware mixture
  of experts.
\newblock In \emph{Proceedings of the IEEE/CVF Conference on Computer Vision
  and Pattern Recognition}, 2021.

\bibitem[Feng et~al.(2020)Feng, You, Chen, Zhang, Zhu, Wu, Wu, and
  Chen]{feng2020kd3a}
Hao-Zhe Feng, Zhaoyang You, Minghao Chen, Tianye Zhang, Minfeng Zhu, Fei Wu,
  Chao Wu, and Wei Chen.
\newblock Kd3a: Unsupervised multi-source decentralized domain adaptation via
  knowledge distillation.
\newblock In \emph{International Conference on Machine Learning}, 2020.

\bibitem[Ge et~al.(2020)Ge, Zhu, Chen, Zhao, and Li]{ge2020selfpaced}
Yixiao Ge, Feng Zhu, Dapeng Chen, Rui Zhao, and Hongsheng Li.
\newblock Self-paced contrastive learning with hybrid memory for domain
  adaptive object re-id.
\newblock In \emph{Advances in Neural Information Processing Systems}, 2020.

\bibitem[Guo et~al.(2020)Guo, Zhu, Zhao, Cao, Lei, and Li]{guo2020learning}
Jianzhu Guo, Xiangyu Zhu, Chenxu Zhao, Dong Cao, Zhen Lei, and Stan~Z Li.
\newblock Learning meta face recognition in unseen domains.
\newblock In \emph{Proceedings of the IEEE/CVF Conference on Computer Vision
  and Pattern Recognition}, 2020.

\bibitem[He et~al.(2016)He, Zhang, Ren, and Sun]{he2016deep}
Kaiming He, Xiangyu Zhang, Shaoqing Ren, and Jian Sun.
\newblock Deep residual learning for image recognition.
\newblock In \emph{Proceedings of the IEEE/CVF Conference on Computer Vision
  and Pattern Recognition}, 2016.

\bibitem[Hermans et~al.(2017)Hermans, Beyer, and Leibe]{hermans2017defense}
Alexander Hermans, Lucas Beyer, and Bastian Leibe.
\newblock In defense of the triplet loss for person re-identification.
\newblock \emph{arXiv preprint arXiv:1703.07737}, 2017.

\bibitem[Ioffe \& Szegedy(2015)Ioffe and Szegedy]{ioffe2015batch}
Sergey Ioffe and Christian Szegedy.
\newblock Batch normalization: Accelerating deep network training by reducing
  internal covariate shift.
\newblock In \emph{International Conference on Machine Learning}, pp.\
  448--456, 2015.

\bibitem[Jin et~al.(2020)Jin, Lan, Zeng, Chen, and Zhang]{jin2020style}
Xin Jin, Cuiling Lan, Wenjun Zeng, Zhibo Chen, and Li~Zhang.
\newblock Style normalization and restitution for generalizable person
  re-identification.
\newblock In \emph{Proceedings of the IEEE/CVF Conference on Computer Vision
  and Pattern Recognition}, 2020.

\bibitem[Kingma \& Welling(2013)Kingma and Welling]{kingma2013auto}
Diederik~P Kingma and Max Welling.
\newblock Auto-encoding variational bayes.
\newblock \emph{arXiv preprint arXiv:1312.6114}, 2013.

\bibitem[Li et~al.(2018)Li, Yang, Song, and Hospedales]{li2018learning}
Da~Li, Yongxin Yang, Yi-Zhe Song, and Timothy~M Hospedales.
\newblock Learning to generalize: Meta-learning for domain generalization.
\newblock In \emph{Proceedings of the AAAI Conference on Artificial
  Intelligence}, 2018.

\bibitem[Li et~al.(2021)Li, He, and Song]{li2021model}
Qinbin Li, Bingsheng He, and Dawn Song.
\newblock Model-contrastive federated learning.
\newblock In \emph{Proceedings of the IEEE/CVF Conference on Computer Vision
  and Pattern Recognition}, 2021.

\bibitem[Li et~al.(2020)Li, Sahu, Zaheer, Sanjabi, Talwalkar, and
  Smith]{li2018federated}
Tian Li, Anit~Kumar Sahu, Manzil Zaheer, Maziar Sanjabi, Ameet Talwalkar, and
  Virginia Smith.
\newblock Federated optimization in heterogeneous networks.
\newblock In \emph{Proceedings of Machine Learning and Systems}, 2020.

\bibitem[Li \& Wang(2013)Li and Wang]{li2013locally}
Wei Li and Xiaogang Wang.
\newblock Locally aligned feature transforms across views.
\newblock In \emph{Proceedings of the IEEE/CVF Conference on Computer Vision
  and Pattern Recognition}, 2013.

\bibitem[Li et~al.(2014)Li, Zhao, Xiao, and Wang]{li2014deepreid}
Wei Li, Rui Zhao, Tong Xiao, and Xiaogang Wang.
\newblock Deepreid: Deep filter pairing neural network for person
  re-identification.
\newblock In \emph{Proceedings of the IEEE/CVF Conference on Computer Vision
  and Pattern Recognition}, 2014.

\bibitem[Liao \& Shao(2020)Liao and Shao]{liao2020interpretable}
Shengcai Liao and Ling Shao.
\newblock Interpretable and generalizable person re-identification with
  query-adaptive convolution and temporal lifting.
\newblock In \emph{European Conference on Computer Vision}, 2020.

\bibitem[Liu et~al.(2021)Liu, Chen, Qin, Dou, and Heng]{liu2021feddg}
Quande Liu, Cheng Chen, Jing Qin, Qi~Dou, and Pheng-Ann Heng.
\newblock Feddg: Federated domain generalization on medical image segmentation
  via episodic learning in continuous frequency space.
\newblock In \emph{Proceedings of the IEEE/CVF Conference on Computer Vision
  and Pattern Recognition}, 2021.

\bibitem[Luo et~al.(2021)Luo, Chen, Hu, Zhang, Liang, and Feng]{luo2021no}
Mi~Luo, Fei Chen, Dapeng Hu, Yifan Zhang, Jian Liang, and Jiashi Feng.
\newblock No fear of heterogeneity: Classifier calibration for federated
  learning with non-iid data.
\newblock In \emph{Advances in Neural Information Processing Systems}, 2021.

\bibitem[McMahan et~al.(2017)McMahan, Moore, Ramage, Hampson, and
  y~Arcas]{mcmahan2017communication}
Brendan McMahan, Eider Moore, Daniel Ramage, Seth Hampson, and Blaise~Aguera
  y~Arcas.
\newblock Communication-efficient learning of deep networks from decentralized
  data.
\newblock In \emph{Artificial intelligence and statistics}, 2017.

\bibitem[Muandet et~al.(2013)Muandet, Balduzzi, and
  Sch{\"o}lkopf]{muandet2013domain}
Krikamol Muandet, David Balduzzi, and Bernhard Sch{\"o}lkopf.
\newblock Domain generalization via invariant feature representation.
\newblock In \emph{International Conference on Machine Learning}, 2013.

\bibitem[Panareda~Busto \& Gall(2017)Panareda~Busto and Gall]{panareda2017open}
Pau Panareda~Busto and Juergen Gall.
\newblock Open set domain adaptation.
\newblock In \emph{Proceedings of the IEEE International Conference on Computer
  Vision}, 2017.

\bibitem[Qiao et~al.(2020)Qiao, Zhao, and Peng]{qiao2020learning}
Fengchun Qiao, Long Zhao, and Xi~Peng.
\newblock Learning to learn single domain generalization.
\newblock In \emph{Proceedings of the IEEE/CVF Conference on Computer Vision
  and Pattern Recognition}, 2020.

\bibitem[Shu et~al.(2021)Shu, Cao, Wang, Wang, and Long]{shu2021open}
Yang Shu, Zhangjie Cao, Chenyu Wang, Jianmin Wang, and Mingsheng Long.
\newblock Open domain generalization with domain-augmented meta-learning.
\newblock In \emph{Proceedings of the IEEE/CVF Conference on Computer Vision
  and Pattern Recognition}, 2021.

\bibitem[Song et~al.(2019)Song, Yang, Song, Xiang, and
  Hospedales]{song2019generalizable}
Jifei Song, Yongxin Yang, Yi-Zhe Song, Tao Xiang, and Timothy~M Hospedales.
\newblock Generalizable person re-identification by domain-invariant mapping
  network.
\newblock In \emph{Proceedings of the IEEE/CVF Conference on Computer Vision
  and Pattern Recognition}, 2019.

\bibitem[Sun et~al.(2018)Sun, Zheng, Yang, Tian, and Wang]{sun2018beyond}
Yifan Sun, Liang Zheng, Yi~Yang, Qi~Tian, and Shengjin Wang.
\newblock Beyond part models: Person retrieval with refined part pooling (and a
  strong convolutional baseline).
\newblock In \emph{European Conference on Computer Vision}, 2018.

\bibitem[Van~der Maaten \& Hinton(2008)Van~der Maaten and
  Hinton]{van2008visualizing}
Laurens Van~der Maaten and Geoffrey Hinton.
\newblock Visualizing data using t-sne.
\newblock \emph{Journal of machine learning research}, 2008.

\bibitem[Wang et~al.(2018)Wang, Yuan, Chen, Li, and Zhou]{wang2018learning}
Guanshuo Wang, Yufeng Yuan, Xiong Chen, Jiwei Li, and Xi~Zhou.
\newblock Learning discriminative features with multiple granularities for
  person re-identification.
\newblock In \emph{Proceedings of the ACM International Conference on
  Multimedia}, 2018.

\bibitem[Wei et~al.(2018)Wei, Zhang, Gao, and Tian]{wei2018person}
Longhui Wei, Shiliang Zhang, Wen Gao, and Qi~Tian.
\newblock Person transfer gan to bridge domain gap for person
  re-identification.
\newblock In \emph{Proceedings of the IEEE/CVF Conference on Computer Vision
  and Pattern Recognition}, 2018.

\bibitem[Wu \& Gong(2021{\natexlab{a}})Wu and Gong]{wu2020decentralised}
Guile Wu and Shaogang Gong.
\newblock Decentralised learning from independent multi-domain labels for
  person re-identification.
\newblock In \emph{Proceedings of the AAAI Conference on Artificial
  Intelligence}, 2021{\natexlab{a}}.

\bibitem[Wu \& Gong(2021{\natexlab{b}})Wu and Gong]{wu2021collaborative}
Guile Wu and Shaogang Gong.
\newblock Collaborative optimization and aggregation for decentralized domain
  generalization and adaptation.
\newblock In \emph{Proceedings of the IEEE/CVF International Conference on
  Computer Vision}, 2021{\natexlab{b}}.

\bibitem[Yang et~al.(2021)Yang, Zhong, Luo, Cai, Lin, Li, and
  Sebe]{yang2021joint}
Fengxiang Yang, Zhun Zhong, Zhiming Luo, Yuanzheng Cai, Yaojin Lin, Shaozi Li,
  and Nicu Sebe.
\newblock Joint noise-tolerant learning and meta camera shift adaptation for
  unsupervised person re-identification.
\newblock In \emph{Proceedings of the IEEE/CVF Conference on Computer Vision
  and Pattern Recognition}, 2021.

\bibitem[Yao et~al.(2022)Yao, Gong, Qi, Cui, Zhu, and Yang]{yao2022federated}
Chun-Han Yao, Boqing Gong, Hang Qi, Yin Cui, Yukun Zhu, and Ming-Hsuan Yang.
\newblock Federated multi-target domain adaptation.
\newblock In \emph{Proceedings of the IEEE/CVF Winter Conference on
  Applications of Computer Vision}, 2022.

\bibitem[Ye et~al.(2021)Ye, Shen, Lin, Xiang, Shao, and Hoi]{pami21reidsurvey}
Mang Ye, Jianbing Shen, Gaojie Lin, Tao Xiang, Ling Shao, and Steven C.~H. Hoi.
\newblock Deep learning for person re-identification: A survey and outlook.
\newblock \emph{IEEE Transactions on Pattern Analysis and Machine
  Intelligence}, 2021.

\bibitem[Yin et~al.(2020)Yin, Molchanov, Alvarez, Li, Mallya, Hoiem, Jha, and
  Kautz]{yin2020dreaming}
Hongxu Yin, Pavlo Molchanov, Jose~M Alvarez, Zhizhong Li, Arun Mallya, Derek
  Hoiem, Niraj~K Jha, and Jan Kautz.
\newblock Dreaming to distill: Data-free knowledge transfer via deepinversion.
\newblock In \emph{Proceedings of the IEEE/CVF Conference on Computer Vision
  and Pattern Recognition}, 2020.

\bibitem[Yoon et~al.(2021)Yoon, Shin, Hwang, and Yang]{yoon2021fedmix}
Tehrim Yoon, Sumin Shin, Sung~Ju Hwang, and Eunho Yang.
\newblock Fedmix: Approximation of mixup under mean augmented federated
  learning.
\newblock In \emph{International Conference on Learning Representation}, 2021.

\bibitem[Zhang et~al.(2017)Zhang, Cisse, Dauphin, and
  Lopez-Paz]{zhang2017mixup}
Hongyi Zhang, Moustapha Cisse, Yann~N Dauphin, and David Lopez-Paz.
\newblock Mixup: Beyond empirical risk minimization.
\newblock In \emph{International Conference on Machine Learning}, 2017.

\bibitem[Zhao et~al.(2018)Zhao, Li, Lai, Suda, Civin, and
  Chandra]{zhao2018federated}
Yue Zhao, Meng Li, Liangzhen Lai, Naveen Suda, Damon Civin, and Vikas Chandra.
\newblock Federated learning with non-iid data.
\newblock \emph{arXiv preprint arXiv:1806.00582}, 2018.

\bibitem[Zhao et~al.(2021)Zhao, Zhong, Yang, Luo, Lin, Li, and
  Sebe]{zhao2021learning}
Yuyang Zhao, Zhun Zhong, Fengxiang Yang, Zhiming Luo, Yaojin Lin, Shaozi Li,
  and Nicu Sebe.
\newblock Learning to generalize unseen domains via memory-based multi-source
  meta-learning for person re-identification.
\newblock In \emph{Proceedings of the IEEE/CVF Conference on Computer Vision
  and Pattern Recognition}, 2021.

\bibitem[Zheng et~al.(2021)Zheng, Liu, He, Mei, Luo, and Zha]{zheng2021group}
Kecheng Zheng, Wu~Liu, Lingxiao He, Tao Mei, Jiebo Luo, and Zheng-Jun Zha.
\newblock Group-aware label transfer for domain adaptive person
  re-identification.
\newblock In \emph{Proceedings of the IEEE/CVF Conference on Computer Vision
  and Pattern Recognition}, 2021.

\bibitem[Zheng et~al.(2015)Zheng, Shen, Tian, Wang, Wang, and
  Tian]{zheng2015scalable}
Liang Zheng, Liyue Shen, Lu~Tian, Shengjin Wang, Jingdong Wang, and Qi~Tian.
\newblock Scalable person re-identification: A benchmark.
\newblock In \emph{Proceedings of the IEEE/CVF International Conference on
  Computer Vision}, 2015.

\bibitem[Zhong et~al.(2019)Zhong, Zheng, Luo, Li, and
  Yang]{zhong2019invariance}
Zhun Zhong, Liang Zheng, Zhiming Luo, Shaozi Li, and Yi~Yang.
\newblock Invariance matters: Exemplar memory for domain adaptive person
  re-identification.
\newblock In \emph{Proceedings of the IEEE/CVF Conference on Computer Vision
  and Pattern Recognition}, 2019.

\bibitem[Zhong et~al.(2020)Zhong, Zheng, Kang, Li, and Yang]{zhong2020random}
Zhun Zhong, Liang Zheng, Guoliang Kang, Shaozi Li, and Yi~Yang.
\newblock Random erasing data augmentation.
\newblock In \emph{Proceedings of the AAAI Conference on Artificial
  Intelligence}, 2020.

\bibitem[Zhou et~al.(2020)Zhou, Yang, Hospedales, and Xiang]{zhou2020learning}
Kaiyang Zhou, Yongxin Yang, Timothy Hospedales, and Tao Xiang.
\newblock Learning to generate novel domains for domain generalization.
\newblock In \emph{European Conference on Computer Vision}, 2020.

\bibitem[Zhuang et~al.(2020)Zhuang, Wen, Zhang, Gan, Yin, Zhou, Zhang, and
  Yi]{zhuang2020performance}
Weiming Zhuang, Yonggang Wen, Xuesen Zhang, Xin Gan, Daiying Yin, Dongzhan
  Zhou, Shuai Zhang, and Shuai Yi.
\newblock Performance optimization of federated person re-identification via
  benchmark analysis.
\newblock In \emph{Proceedings of the ACM International Conference on
  Multimedia}, 2020.

\end{thebibliography}
